\documentclass{article}

\usepackage{microtype}
\usepackage{graphicx}
\usepackage{subfigure}
\usepackage{booktabs} 

\usepackage{amsmath,amsfonts,bm}

\def\eqref#1{equation~\ref{#1}}

\def\1{\bm{1}}

\def\vh{{\bm{h}}}

\DeclareMathAlphabet{\mathsfit}{\encodingdefault}{\sfdefault}{m}{sl}
\SetMathAlphabet{\mathsfit}{bold}{\encodingdefault}{\sfdefault}{bx}{n}

\def\sV{{\mathbb{V}}}

\usepackage{hyperref}

\usepackage[accepted]{icml2025}

\usepackage{amsmath}
\usepackage{amssymb}
\usepackage{mathtools}
\usepackage{amsthm}

\usepackage[capitalize,noabbrev]{cleveref}

\theoremstyle{plain}
\newtheorem{theorem}{Theorem}[section]
\newtheorem{proposition}[theorem]{Proposition}

\theoremstyle{definition}

\theoremstyle{remark}

\usepackage[textsize=tiny]{todonotes}
\usepackage{hyperref}
\usepackage{url}
\usepackage{graphicx}
\usepackage{wrapfig}
\usepackage{booktabs}
\usepackage{multirow}
\usepackage{enumitem}

\icmltitlerunning{Equivariant Random GNN}

\begin{document}

\twocolumn[
\icmltitle{Using Random Noise Equivariantly to Boost Graph Neural Networks Universally}



\icmlsetsymbol{equal}{*}

\begin{icmlauthorlist}
\icmlauthor{Xiyuan Wang}{iai}
\icmlauthor{Muhan Zhang}{iai}
\end{icmlauthorlist}

\icmlaffiliation{iai}{Institute for Artificial Intelligence, Peking University}

\icmlcorrespondingauthor{Muhan Zhang}{wangxiyuan@pku.edu.cn, muhan@pku.edu.cn}

\icmlkeywords{Machine Learning, ICML}

\vskip 0.3in
]



\printAffiliationsAndNotice{}  

\begin{abstract}
Recent advances in Graph Neural Networks (GNNs) have explored the potential of random noise as an input feature to enhance expressivity across diverse tasks. However, naively incorporating noise can degrade performance, while architectures tailored to exploit noise for specific tasks excel yet lack broad applicability. This paper tackles these issues by laying down a theoretical framework that elucidates the increased sample complexity when introducing random noise into GNNs without careful design. We further propose Equivariant Noise GNN (ENGNN), a novel architecture that harnesses the symmetrical properties of noise to mitigate sample complexity and bolster generalization. Our experiments demonstrate that using noise equivariantly significantly enhances performance on node-level, link-level, subgraph, and graph-level tasks and achieves comparable performance to models designed for specific tasks, thereby offering a general method to boost expressivity across various graph tasks.
\end{abstract}

\section{Introduction}
Graph Neural Networks (GNNs) have emerged as powerful tools for graph representation learning, delivering state-of-the-art performance in applications such as natural language processing~\citep{intro_nlp}, bioinformatics~\citep{intro_bio}, and social network analysis~\citep{intro_soc}. However, even the most popular GNN architectures, including Message Passing Neural Networks (MPNNs)~\citep{MPNN}, face limitations in expressivity when tackling complex node-, link-, and graph-level tasks~\citep{LRGB,oversmoothing1,oversquashing1,SEAL,labelingtrick}. These limitations hinder their ability to fully capture intricate graph structures.

Efforts to improve the expressivity of GNNs generally fall into two categories. The first focuses on enhancing GNN architectures. This includes improving message aggregation methods~\citep{GIN, PNA, JacobiConv}, leveraging high-order tensors to capture complex structures~\citep{PyGHO, PPGN, IGN, NGNN, I2GNN, ESAN, OSAN, SUN, GNNAK, SSWL}, and incorporating transformers to boost global information capacity~\citep{GT-graphit, GT-SAN, GT-graphtrans, firstGT, GT-graphormer, GT-Exphormer, GT-GPS, GT-DeepSet, GraphAsSet, firstGT}. The second category introduces auxiliary input features for GNNs, such as hand-crafted positional and structural encodings for graph-level tasks~\citep{RWSE, DE, EquiStableEnc, SignBasis-inv, huang2023stability, LCSE, DE}, pairwise neighborhood overlap features for link prediction~\citep{NCN, BUDDY, NeoGNN}, and random noise for various tasks~\citep{rGIN, GNNRNI, mplp, randomeigen}.

\begin{figure}[t]
    \centering
    \vskip -0.1in
    \includegraphics[width=0.95\linewidth]{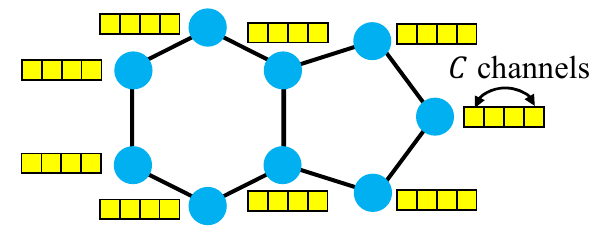}
    \vskip -0.1in
    \caption{Example of GNNs with noise as auxiliary input. Each node has $C$ channel noise. }
    \label{fig:noiseexp}
    \vskip -0.2in
\end{figure} 
Among these approaches, random noise-based methods are particularly appealing. As illustrated in Figure~\ref{fig:noiseexp}, these methods assign each node a multi-channel random vector as an additional input feature. This avoids the complexity of intricate architectures while remaining compatible with various architectural improvements. Furthermore, unlike other auxiliary features, random noise is task-agnostic and can provably increase GNN expressivity~\citep{rGIN}.

Despite these advantages—strong expressivity, low complexity, and compatibility with other designs—random noise is not widely used in graph tasks. One reason is that directly feeding noise to GNN leads to poor generalization. Early works that directly incorporated noise~\citep{rGIN, GNNRNI} primarily focused on synthetic datasets, where expressivity dominates performance and generalization is less important. On real-world datasets, however, the poor generalization induced by noise often outweighs the performance gains from increased expressivity, resulting in overall performance degradation. 

Recently, methods like MPLP~\citep{mplp} and LLwLC~\citep{randomeigen} have successfully applied noise to real-world link prediction tasks. These approaches carefully design linear GNNs to produce interpretable and stable output from noise: MPLP approximates the overlap of two nodes' neighborhoods, and LLwLC approximates Laplacian eigenvectors. These designs reduce variance and ensure convergence as noise dimension increases. However, because these methods rely on specific heuristics for link prediction, they lack flexibility and cannot generalize to other tasks.

In this work, we aim to design a general method for leveraging noise features to boost GNN performance. We first analyze the generalization of GNNs with noise input and find that introducing noise significantly increases sample complexity, which scales with the noise space ``volume''. Intuitively, ordinary GNNs treat the same graph with different noise as distinct inputs, requiring the model to learn each graph repeatedly with different noise, thus multiplying the sample complexity, which explains the poor generalization observed with noise inputs. 

Methods like MPLP and LLwLC mitigate this issue by mapping the same graph with different noise to similar representations, effectively reducing the points needed to cover the noise space and alleviating generalization problems. Building on this insight, we propose that the key to minimizing the adverse effects of noise is to prevent models from treating the same graph with different noise configurations as entirely distinct graphs. To achieve this, we design models that are invariant to specific group operations on the noise, thereby reducing sample complexity.

Based on theoretical analysis, we propose \textbf{Equivariant Noise GNN (ENGNN)}, an architecture featuring a channel-permutation-equivariant aggregator for processing noise features. This design preserves universal expressivity while improving generalization. ENGNN demonstrates consistent performance gain compared with ordinary MPNNs across node-, link-, subgraph-, and graph-level tasks, verifying its effectiveness for general graph problems.


\section{Preliminaries}

For a matrix $ Z \in \mathbb{R}^{a \times b} $, let $ Z_i \in \mathbb{R}^b $ denote the $ i $-th row (as a column vector), $ Z_{:, j} \in \mathbb{R}^a $ denote the $ j $-th column, and $ Z_{ij} \in \mathbb{R} $ denote the element at the $ (i, j) $-th position. An \textit{undirected graph} is represented as $ \mathcal{G} = (V, E, X) $, where $ V = \{1, 2, 3, \dots, n\} $ is the set of $ n $ nodes, $ E \subseteq V \times V $ is the set of edges, and $ X \in \mathbb{R}^{n \times d} $ is the node feature matrix, with the $ v $-th row $ X_v $ representing the features of node $ v $. The edge set $ E $ can also be expressed using the adjacency matrix $ A \in \mathbb{R}^{n \times n} $, where $ A_{uv} $ is 1 if the edge $ (u, v) $ exists, and 0 otherwise. A graph $\mathcal{G}$ can therefore be represented by the tuples $ (V, A, X) $ or simply $ (A, X) $. 

\paragraph{Message Passing Neural Network (MPNN)~\citep{MPNN}} MPNN is a popular GNN framework. It consists of multiple message-passing layers, where the $ k $-th layer is:
\begin{multline}
\vh_v^{(k)} = U^{(k)}\large(\vh_v^{(k-1)}, \text{AGG}(\\
\{M^{(k)}(\vh_u^{(k-1)}) \mid u \in V, (u, v) \in E\})\large),
\end{multline}
where $ \vh_v^{(k)} $ is the representation of node $ v $ at the $ k $-th layer, $ U^{(k)} $ and $ M^{(k)} $ are functions such as Multi-Layer Perceptrons (MLPs), and $ \text{AGG} $ is an aggregation function like sum or max. The initial node representation $ \vh_v^{(0)} $ is the node feature $ X_v $. Each layer aggregates information from neighbors to update the center node's representation. 

\paragraph{Equivariance and Invariance} Given a function $h: \mathcal{X} \to \mathcal{Y}$ and a set of operator $T$ acting on $\mathcal{X}$ and $\mathcal{Y}$ through operation $\star$, $h$ is \textit{$T$-invariant} if $h(t \star x) = h(x), \quad \forall x \in \mathcal{X}, t \in T$, and \textit{$T$-equivariant} if $h(t \star x) = t \star h(x), \quad \forall x \in \mathcal{X}, t \in T$. 

Most graph properties are invariant to the ordering of nodes, implying invariance to node permutations. Considered as functions of the input, representations can be classified as equivariant or invariant by whether they preserve these properties under the group operations on input.

\section{Generalization}
In this section, we use sample complexity in PAC (Probably Approximately Correct) theory to explore why GNNs with noise inputs tend to have poor generalization performance and why keeping invariance to noise can solve this problem. All proofs for the theory in this section are in Appendix~\ref{app:gen_proof}.

\paragraph{Notations}

Given graph $G = (A, X) \in \mathcal{G}$ with $n$ nodes, the noise is represented by $Z \in \mathcal{Z} = \mathbb{R}^{n \times C}$, where $C$ is the number of noise channels per node. The prediction target is in a compact set $\mathcal{Y} \subseteq \mathbb{R}^d$. The \textit{hypothesis class} $H$ is set of functions mapping a pair of graph and noise to prediction $h: \mathcal{G} \times \mathcal{Z} \to \mathcal{Y}$. The learning task involves a \textit{loss function} $l: Y \times Y \to [0, 1]$, which we assume to be Lipschitz continuous with constant $C_l$. The learning algorithm, denoted by $\text{alg}$, maps a finite set of training data points to a hypothesis in $H$. 

The \textit{sample complexity} of an algorithm refers to the number of training samples needed to achieve a certain level of generalization, given by $\epsilon$ (the error bound) and $\delta$ (the probability of failure). A higher sample complexity generally means poorer generalization. Intuitively, the larger the input space, the more data points are needed to achieve a good generalization. To measure the size of a space, we introduce the concept of \textit{covering numbers}. Let $(U, \rho)$ be a \textit{pseudo-metric} space, where $\rho$ measures the distance between elements in $U$. In our case, we define $\rho_G$, $\rho_Z$, and $\rho_Y$ as the pseudo-metric functions on the graph space $\mathcal{G}$, noise space $\mathcal{Z}$, and label space $\mathcal{Y}$, respectively. The covering number $N(U, \rho, r)$ is the minimal number of points required to cover the entire space $U$, such that for each point $u \in U$, there exists $u_j$ in the selected points with $\rho(u, u_j) \leq r$. 

\paragraph{Symmetry in Noise Space}

We assume that the model is invariant to some transformations on noise. Let $T = \{ t: \mathcal{Z} \to \mathcal{Z} \mid \forall G \in \mathcal{G}, Z \in \mathcal{Z}, h(G, t(Z)) = h(G, Z) \}$ represent \textit{the set of noise transformations that leave the output unchanged}. If the model is not designed to maintain invariance under certain operations, we can set $T$ to a set with identity mapping only.

When the model is invariant to operators in $T$, two noise points that are distant from each other in the original space may be transformed to two closer points, which allows the model to treat them similarly and reduce the sample complexity. We define the pseudo-metric function induced by the transformation set $T$ as:
\begin{equation}
\rho_{Z, T}(Z_1, Z_2) = \inf_{t_1 \in T, t_2 \in T} \rho_Z(t_1(Z_1), t_2(Z_2)).
\end{equation}
This transformed metric reduces the distance between noise points, so fewer samples are needed to cover the noise space.

\paragraph{Sample Complexity for Models with Graph and Noise Inputs}

Now, we consider the sample complexity for a model that takes both graph and noise inputs. Let $T$ be the set of transformations applied to the noise. If the hypothesis class consists only of $T$-invariant functions, and if the learning algorithm is also $T$-invariant (like empirical loss minimization), the sample complexity can be reduced. The following theorem provides a bound on the sample complexity:
\begin{theorem}
Let $T \subseteq \{ t: \mathcal{Z} \to \mathcal{Z} \}$. Assume that all hypotheses $h \in H$ are $C_G$-Lipschitz with respect to the graph space $\mathcal{G}$ and $C_Z$-Lipschitz with respect to the noise space $\mathcal{Z}$, and that $h$ is $T$-invariant. If the algorithm is also $T$-invariant, then the sample complexity is:

\begin{equation}\label{equ:samplecomplex}
O\left( \frac{1}{\epsilon^2} N_G N_Z \ln N_Y + \frac{1}{\epsilon^2} \ln \frac{1}{\delta} \right),
\end{equation}

\begin{itemize}[itemsep=2pt,topsep=-2pt,parsep=0pt,leftmargin=10pt]
    \item where $N_G\!=\!N(\mathcal{G}, \rho_G, \frac{\delta}{12C_lC_G}), N_Y\!=\!N(\mathcal{Y}, \rho_Y, \frac{\delta}{12C_l})$ are the covering numbers for the graph space $\mathcal{G}$ and the label space $\mathcal{Y}$, respectively.
    \item $N_Z=N(\mathcal{Z}, \rho_{Z, T}, \frac{\delta}{12C_lC_Z})$ is the covering number for noise space $\mathcal{Z}$ with pseudometrics induced by $T$,
    \item $\epsilon, \delta$ are the error bound and failure probability,
\end{itemize}
\end{theorem}

In Equation~\ref{equ:samplecomplex}, the second term $\frac{1}{\epsilon^2} \ln \frac{1}{\delta}$ arises from concentration inequalities and is independent of the task. The first term depends on the covering numbers of the graph space $\mathcal{G}$ and the noise space $\mathcal{Z}$. When there is no noise (i.e., $\mathcal{Z}$ is a constant matrix), $N_Z=1$. When noise is introduced, the sample complexity increases because $N_Z>1$ scales the first term. This helps explain why noise often results in poorer generalization: as the model becomes more sensitive to noise variations, the covering number for the noise space increases, which raises the sample complexity. However, the additional expressiveness from the noise can still help improve the test performance by reducing the training loss.

\paragraph{Reducing Sample Complexity Through Symmetry}

When we apply symmetries to the noise space, the sample complexity can be reduced. For instance, if we consider a set $T_1 \subseteq T_2$, the covering number for $T_1$ will be larger or equal to that for $T_2$, as adding more operations to $T$ reduces the distance between points in $\mathcal{Z}$. Formally:

\begin{proposition}
If $T_1 \subseteq T_2$, then:

\begin{equation}
N(\mathcal{Z}, \rho_{Z, T_1}, \tau) \ge N(\mathcal{Z}, \rho_{Z, T_2}, \tau).
\end{equation}

\end{proposition}

Thus, the more operations we include in $T$, the lower the sample complexity. For example, if we consider the permutations of the noise channels, we get the following result:

\begin{proposition}
If $\mathcal{Z} = [0, 1]^{n \times C}$, and all functions in $H$ are invariant to permutations of the $C$-channel noise, then:
\begin{equation}
\frac{N(\mathcal{Z}, \rho_{Z, T}, \frac{\delta}{12C_lC_Z})}{N(\mathcal{Z}, \rho_Z, \frac{\delta}{12C_lC_Z})} = \frac{1}{C!}.
\end{equation}
\end{proposition}

This ratio can be quite large, significantly reducing the sample complexity when the model is invariant to noise permutations. Previous methods can be interpreted within this framework: \citet{mplp} achieve invariance to orthogonal transformations on noise, while \citet{randomeigen} achieve invariance to rotations of noise about eigenvector axes. Orthogonal group is ``larger'' than permutation group and can further reduce sample complexity, but their designs constrain them to linear GNNs and task-specific properties, whereas our theory framework allows more flexible models. 

\section{Equivariant Noise Graph Neural Network}
This section propose our Equivariant Noise Graph Neural Network (ENGNN). All proofs are in Appendix~\ref{app:exp_proof}. ENGNN is designed with two key objectives:  1. To achieve strong expressivity.  2. To preserve as much symmetry as possible, which enhances the model's generalization capabilities. They can sometimes conflict: Directly feeding noise into standard MPNNs can achieve universal expressivity~\citep{rGIN} but at the cost of completely ignoring symmetry. Conversely, maintaining invariance to larger symmetry groups (e.g., orthogonal groups) requires high-order representations for high expressivity~\cite{TFNuniversality}, which introduces significant computational overhead and is hard to implement. \textbf{To balance these trade-offs, our approach focuses on preserving invariance to the permutation group of noise channels, enabling us to design an architecture that is both expressive and efficient.}

In ENGNN, each node $i$ is represented by two components: An \textit{equivariant representation} $Z_i \in \mathbb{R}^{L \times C}$, which transforms with permutations of the noise channels in the input noise $Z_i^{(0)} \in \mathbb{R}^C$, and an \textit{invariant representation} $X_i \in \mathbb{R}^d$, which remains unchanged under noise permutations. Here, $d$ and $L$ denote hidden dimensions, and $C$ is the number of noise channels. Initially, the equivariant representation $Z^{(0)}$ is set to the input noise, while the invariant representation $X^{(0)}$ is initialized using the node features.

With MPNN architecture, MPNN passes both invariant and equivariant messages between nodes. After several message-passing layers, the updated node representations are pooled to produce representations for the target node sets. Therefore, we first build an injective equivariant aggregator, which processes pairs of invariant and equivariant representations. This aggregator serves as the basis for constructing both the message-passing and pooling layers.

\subsection{Equivariant Aggregator}
To process both invariant and equivariant features while respecting symmetry, we design a specialized equivariant aggregator based on the \textit{DeepSet} model~\citep{DeepSet}, inspired by prior work~\citep{SymDeepSet, NewDeepSet}. DeepSet aggregates invariant features $x_1,...,x_k\in \mathbb{R}^d$ to produce representation $z\in \mathbb{R}^{d'}$ as follows,
\begin{equation}
    z = \text{MLP}_2\left(\sum_{i=1}^k \text{MLP}_1(x_i)\right),
\end{equation}
where $d$ and $d'$ denote the hidden dimensions, and $\text{MLP}_1, \text{MLP}_2$ represent two MLPs. Initially, DeepSet transforms each element in the set individually through $\text{MLP}_1$, then sums all transformed elements to produce set statistics, which are further transformed by $\text{MLP}_2$.

Generalizing DeepSet, our aggregator transforms pairs of invariant and equivariant features while ensuring consistency under permutations of nodes and noise channels. The equivariant aggregator takes a set of invariant features $X \in \mathbb{R}^{k \times d}$ and equivariant noise features $Z \in \mathbb{R}^{k \times L \times C}$, where $k$ is the number of nodes or elements, $d$ and $L$ are hidden dimensions for invariant and equivariant features, and $C$ is the number of noise channels. The goal is to express a function $\text{AGGR}: \mathbb{R}^{k \times d} \times \mathbb{R}^{k \times L \times C} \to \mathbb{R}^{k \times d'} \times \mathbb{R}^{k \times L' \times C}$ that respects permutations of nodes (denoted as permutation group $S_k$) and noise channels (denoted as permutation group $S_C$). 

The aggregator follows four main steps:
\begin{enumerate}[itemsep=2pt,topsep=-2pt,parsep=0pt,leftmargin=10pt]
    \item \textbf{Identifying Noise Channels}:  
   To distinguish noise channels, we apply a DeepSet model $\psi: \mathbb{R}^{k \times L} \to \mathbb{R}^{L_0}$, invariant to node permutations. This ensures that each noise channel is uniquely identified. For instance, when the noise features $Z_{:, :, i}$ and $Z_{:, :, j}$ have distinct sets of elements, $\psi(Z_{:, :, i}) \neq \psi(Z_{:, :, j})$. By concatenating these identifiers with the original noise features, we obtain:
   \begin{equation}
   Z^1 \in \mathbb{R}^{k \times (L+L_0) \times C}, \quad Z^1_{i, :, k} = Z_{i, :, k} \Vert \psi(Z_{:, :, k}),
   \end{equation}
   where $\Vert$ denotes concatenation.
   \item \textbf{Node Encoding}:  
   To combine noise and invariant features, we apply another DeepSet model $\phi: \mathbb{R}^{(L+L_0) \times C} \to \mathbb{R}^{L_1}$ invariant to noise channel permutations. 
   Concatenating the encoded noise with the original invariant features yields updated node features:
   \begin{equation}
   X^0 \in \mathbb{R}^{k \times (d+L_1)}, \quad X^0_i = \phi(Z^1_i) \Vert X_i.
   \end{equation}
   \item \textbf{Set Encoding}:  
   To aggregate information across all nodes, we use a DeepSet model $\varphi: \mathbb{R}^{k \times (d+L_1)} \to \mathbb{R}^{d_1}$ invariant to node permutations. The resulting aggregated feature combines information from all nodes in the set:
   \begin{equation}
   X^1 \in \mathbb{R}^{d_1}, \quad X^1 = \varphi(X^0).
   \end{equation}
   \item  \textbf{Generating Equivariant Outputs}:  
   Finally, the aggregated features are transformed using two MLPs, $g$ and $h$, to produce updated invariant and equivariant features:
   \begin{equation}
   X^2 \in \mathbb{R}^{k \times d'}, \quad X^2_i = g(X^1 \Vert X^0_i),
   \end{equation}
   \begin{equation}
   Z^2 \in \mathbb{R}^{k \times L' \times C}, \quad Z^2_{i, :, j} = h(X^1 \Vert X^0_i \Vert Z^1_{i, :, j}).
   \end{equation}
\end{enumerate}

Each step of the aggregator is efficient, with a time and space complexity of $\Theta(k)$. 

Let $\text{AGGR}$ denote our aggregator. $\text{AGGR}$ guarantees equivariance to node and noise channel permutations. Formally:
\begin{proposition}\label{prop:aggr_equiv}
For any parameterization of $\psi, \phi, \varphi, g, h$, features $X \in \mathbb{R}^{k \times d}$, noise features $Z \in \mathbb{R}^{k \times L \times C}$, and permutations $P_1 \in S_k$ for node, $P_2 \in S_C$ for noise channel, if $X', Z' = \text{AGGR}(X, Z)$, then:
\begin{equation}
\!\!P_1(X'), P_2(P_1(Z'))=\text{AGGR}(P_1(X),P_2(P_1(Z))).
\end{equation}
\end{proposition}
Under mild conditions, $\text{AGGR}$ can approximate any equivariant continuous function. Formally:
\begin{proposition}\label{prop:aggr_exp}
Given a compact set $U \subseteq \mathbb{R}^{k \times d} \times \mathbb{R}^{k \times L \times C}$ that for all each channel of noise has a different elements multiset, $\text{AGGR}$ is a universal approximator of continuous $S_k \times S_C$-equivariant functions on $U$.
\end{proposition}

This universality ensures that $\text{AGGR}$ can handle various tasks. Additionally, the condition that different noise channels have distinct elements is straightforward to achieve, as we can sample noise until condition is fulfilled.

\subsection{Architecture}

Building on the expressive aggregator, we develop two components: message-passing layers and pooling layers.

\paragraph{Message-Passing Layer}In the message-passing layer, each node $u$ updates its representation based on its neighbors $N(u)$. To avoid redundant computations, channel indicators and node encodings are shared across different neighbors. The process is as follows:
\begin{enumerate}[itemsep=2pt,topsep=-2pt,parsep=0pt,leftmargin=10pt]
    \item \textbf{Identifying Noise Channels}:  
   Compute $Z^1 \in \mathbb{R}^{n \times L + L_0 \times C}$ with DeepSet model $\psi$, where $Z^1_{i, j, k} = Z_{i, j, k} \Vert \psi(Z_{:, :, k})$. This step uniquely encodes noise channels.
   \item \textbf{Node Encoding}: Using a DeepSet model $\phi$, compute $x^0 \in \mathbb{R}^{n \times d_0}$, where $x^0_i = \phi(Z^1_i) \Vert x_i$. This combines noise information with node features.
   \item \textbf{Set Encoding}: Aggregate each node’s neighbor representations with a DeepSet model $\varphi$:  
   \begin{equation}
   x^1 \in \mathbb{R}^{n \times d_1}, \quad x^1_i = \varphi(\{x^0_j \mid j \in N(i)\}).
   \end{equation}
   \item \textbf{Generating Equivariant Outputs}: Compute the final representations using MLPs $g$ and $h$:  
   \begin{equation}
   x^2 \in \mathbb{R}^{n \times d'}, \quad x^2_i = g(x^1 \Vert x^0_i),
   \end{equation}
   \begin{equation}
   Z^2 \in \mathbb{R}^{n \times L' \times C}, \quad Z^2_{i,:,j} = h(x^1 \Vert x^0_i \Vert Z^1_{i,:,j}).
   \end{equation}
\end{enumerate}

The overall complexity of the message-passing layer is $\Theta(m)$, where $m$ is the number of edges.

\paragraph{Pooling Layer}

After several message-passing layers, node representations are pooled to generate a graph-level representation:

\begin{equation}
X', Z' = \text{AGGR}(X, Z), \quad h_G = \sum_{i \in V} X'_i.
\end{equation}

The overall time and space complexity of our model is $O(|E|)$, where $|E|$ is the number of input edges. Our model keeps the same complexity as ordinary MPNNs.

\subsection{Expressivity}

Our model is designed to effectively distinguish between different graph structures while treating isomorphic graphs equivalently. The following theoretical results demonstrate the model’s expressive power:

\begin{theorem}\label{thm:graph_exp}
Let $\text{ENGNN}(G, Z)$ denote the output of our model with graph $G$ and noise $Z$ as input. Given a noise space $\mathcal{Z}$ where each noise channel has a distinct set of elements and each node is assigned unique noise, a node-permutation-invariant noise distribution, if our model has at least one message passing layer and one pooling layer: 
\begin{itemize}[itemsep=2pt,topsep=-2pt,parsep=0pt,leftmargin=10pt]
    \item For any two isomorphic graphs $G$ and $H$, 
  \begin{equation}
  \mathbb{E}_{Z \in \mathcal{Z}} \text{ENGNN}(G,Z) = \mathbb{E}_{Z \in \mathcal{Z}} \text{ENGNN}(H,Z).
  \end{equation}
  \item For any non-isomorphic graphs $G$ and $H$, there exists a parameterization such that for all $Z_1, Z_2\in \mathcal Z$:
  \begin{equation}
  \text{ENGNN}(G,Z_1) \neq \text{ENGNN}(H,Z_2).
  \end{equation}
\end{itemize}
\end{theorem}
This result guarantees that our model produces outputs with identical expectation for graphs that are isomorphic while being expressive enough to differentiate between graphs with different structures. The use of noise ensures that the model captures subtle graph differences, and the averaging over noise space helps maintain this robustness. However, directly computing these expectations may be computationally expensive, and outputs for isomorphic graphs could vary slightly under different noise realizations.

To generalize this expressivity to other \textbf{graph tasks, such as representing nodes, links, or subgraphs}, we propose the following method to produce representations of node subsets in a graph. Given a node subset $U$ (representing a node, link, or subgraph), the process is:
\begin{enumerate}[itemsep=2pt,topsep=-2pt,parsep=0pt,leftmargin=10pt]
    \item Compute the global representation:
   \begin{equation}
   X_G, Z_G = \text{AGGR}(\{(X_i, Z_i) \mid i \in V\}),
   \end{equation}
   where $X_G$ and $Z_G$ represent the invariant and equivariant aggregated features from the whole graph.
   \item Compute the subset-specific representation:
   \begin{equation}
   X_U, Z_U = \text{AGGR}(\{(X_i, Z_i) \mid i \in U\}),
   \end{equation}
   which aggregates features from the nodes in subset $U$.
   \item Concatenate the subset and global representations, and use an MLP to produce the final output:
   \begin{equation}
   h_U = \text{MLP}(X_U \Vert X_G \Vert \text{DeepSet}(Z_U \Vert Z_G)),
   \end{equation}
   where $\text{DeepSet}$ operates along the $L$-dimension, and $v_U \Vert v_G$ denotes concatenation along the $d$-dimension.
\end{enumerate}

\begin{theorem}\label{thm:subgraph_exp}
Let $\text{ENGNN}(U, G, Z)$ denote the output of our model with node subset $U$, graph $G$, and noise $Z$ as input. Given a noise space $\mathcal{Z}$ where each noise channel has a distinct set of elements and each node is assigned unique noise, a node-permutation-invariant noise distribution, if our model has at least one message passing layer and one pooling layer: 
\begin{itemize}[itemsep=2pt,topsep=-2pt,parsep=0pt,leftmargin=10pt]
    \item For any two isomorphic graphs with node subsets $(G, U_G)$ and $(H, U_H)$, the model ensures that:
\begin{small}
\begin{equation}
  \mathbb{E}_{Z} \text{ENGNN}(U_G, G, Z) = \mathbb{E}_{Z} \text{ENGNN}(U_H, H, Z).
  \end{equation}
\end{small}
  \item For any non-isomorphic graphs with node subsets $(G, U_G)$ and $(H, U_H)$, there exists a parameterization such that for all $Z_1, Z_2\in \mathcal Z$:
  \begin{equation}
  \text{ENGNN}(U_G, G,Z_1) \neq \text{ENGNN}(U_H, H,Z_2).
  \end{equation}
\end{itemize}
Here, two graphs with node subsets $(G, U_G)$ and $(H, U_H)$ are isomorphic iff there exists an node permutation $P$ such that $P(G)=H$ and $P(U_G)=P(U_H)$.
\end{theorem}
 
This result extends the model’s ability to differentiate graph structures to the level of individual nodes, links, or subgraphs. Isomorphic graphs and their corresponding subsets are guaranteed to produce identical outputs, while non-isomorphic graphs or subsets can be distinguished through proper parameterization. This makes the model versatile for a wide range of graph-level and local-level tasks.

\begin{table*}
\centering
\caption{Results on node classification datasets: Mean accuracy (\%) $\pm$ standard variation.}\label{tab::node}
\begin{center}
\begin{small}
\begin{sc}
\resizebox{1.0\textwidth}{!}{
\setlength{\tabcolsep}{1mm}
\begin{tabular}{lccccc|ccc}
\toprule
Datasets & GCN & APPNP & ChebyNet & GPRGNN & BernNet & ENGNN & MPNN & NMPNN \\
\midrule
        Cora & $87.14_{\pm 1.01}$ & $88.14_{\pm 0.73}$ & $86.67_{\pm 0.82}$ & $88.57_{\pm 0.69}$ & $88.52_{\pm 0.95}$ & $\mathbf{88.85_{\pm0.96}}$ & $87.36_{\pm0.52}$ & $20.11_{\pm2.01}$ \\ 
        Citeseer & $79.86_{\pm 0.67}$ & $\mathbf{80.47_{\pm 0.74}}$ & $79.11_{\pm 0.75}$ & $80.12_{\pm 0.83}$ & $80.09_{\pm 0.79}$ & $79.97_{\pm0.79}$ & $79.62_{\pm0.75}$ & $20.80_{\pm2.63}$ \\ 
        Pubmed & $86.74_{\pm 0.27}$ & $88.12_{\pm 0.31}$ & $87.95_{\pm 0.28}$ & $88.46_{\pm 0.33}$ & $88.48_{\pm 0.41}$ & $\mathbf{89.79_{\pm0.64}}$ & $89.53_{\pm0.29}$ & $69.28_{\pm3.14}$ \\ 
        Computers & $83.32_{\pm 0.33}$ & $85.32_{\pm 0.37}$ & $87.54_{\pm 0.43}$ & $86.85_{\pm 0.25}$ & $87.64_{\pm 0.44}$ & $\mathbf{90.48_{\pm0.31}}$ & $89.53_{\pm0.83}$ & $66.42_{\pm1.39}$ \\ 
        Photo & $88.26_{\pm 0.73}$ & $88.51_{\pm 0.31}$ & $93.77_{\pm 0.32}$ & $93.85_{\pm 0.28}$ & $93.63_{\pm 0.35}$ & $\mathbf{95.24_{\pm0.58}}$ & $94.74_{\pm0.25}$ & $65.12_{\pm1.95}$ \\ 
        Chameleon & $59.61_{\pm 2.21}$ & $51.84_{\pm 1.82}$ & $59.28_{\pm 1.25}$ & $67.28_{\pm 1.09}$ & $68.29_{\pm 1.58}$ & $\mathbf{71.40_{\pm1.29}}$ & $67.18_{\pm1.07}$ & $41.25_{\pm1.38}$ \\ 
        Actor & $33.23_{\pm 1.16}$ & $39.66_{\pm 0.55}$ & $37.61_{\pm 0.89}$ & $39.92_{\pm 0.67}$ & $\mathbf{41.79_{\pm 1.01}}$ & $40.64_{\pm0.67}$ & $40.41_{\pm1.53}$ & $23.73_{\pm2.36}$ \\ 
        Squirrel & $46.78_{\pm 0.87}$ & $34.71_{\pm 0.57}$ & $40.55_{\pm 0.42}$ & $50.15_{\pm 1.92}$ & $51.35_{\pm 0.73}$ & $\mathbf{52.77_{\pm1.43}}$ & $51.99_{\pm1.78}$ & $38.25_{\pm1.04}$ \\ 
\bottomrule
\end{tabular}
}
\end{sc}
\end{small}
\end{center}
\end{table*}
\begin{table*}[t]
    \centering
\vskip -0.15in
    \caption{Results on link prediction benchmarks. The format is average score $\pm$ standard deviation. OOM means out of GPU memory.}\label{tab::link}
\vskip 0.1in
\setlength{\tabcolsep}{2pt}
    \begin{tabular}{lccccccc}
    \toprule
~ & {Cora} & {Citeseer} & {Pubmed} & {Collab} & {PPA} & {DDI} \\ 
        Metric & HR@100 & HR@100 & HR@100 & HR@50 & HR@100 & HR@20 \\ 
\midrule
        {CN} & $33.92 {\scriptstyle \pm 0.46}$ & $29.79 {\scriptstyle \pm 0.90}$ & $23.13 {\scriptstyle \pm 0.15}$ & $56.44 {\scriptstyle \pm 0.00}$ & $27.65 {\scriptstyle \pm 0.00}$ & $17.73 {\scriptstyle \pm 0.00}$ \\ 
        {AA} & $39.85 {\scriptstyle \pm 1.34}$ & $35.19 {\scriptstyle \pm 1.33}$ & $27.38 {\scriptstyle \pm 0.11}$ & $64.35 {\scriptstyle \pm 0.00}$ & $32.45 {\scriptstyle \pm 0.00}$ & $18.61 {\scriptstyle \pm 0.00}$ \\ 
        {RA} & $41.07 {\scriptstyle \pm 0.48}$ & $33.56 {\scriptstyle \pm 0.17}$ & $27.03 {\scriptstyle \pm 0.35}$ & $64.00 {\scriptstyle \pm 0.00}$ & $49.33{\scriptstyle \pm 0.00}$ & $27.60 {\scriptstyle \pm 0.00}$ \\ 
        {GCN} & $66.79 {\scriptstyle \pm 1.65}$ & $67.08 {\scriptstyle \pm 2.94}$ & $53.02 {\scriptstyle \pm 1.39}$ & $44.75 {\scriptstyle \pm 1.07}$ & $18.67 {\scriptstyle \pm 1.32}$ & $37.07 {\scriptstyle \pm 5.07}$ \\ 
        {SAGE} & $55.02 {\scriptstyle \pm 4.03}$ & $57.01 {\scriptstyle \pm 3.74}$ & $39.66 {\scriptstyle \pm 0.72}$ & $48.10 {\scriptstyle \pm 0.81}$ & $16.55 {\scriptstyle \pm 2.40}$ & $53.90 {\scriptstyle \pm 4.74}$ \\ 
        {SEAL} & $81.71{\scriptstyle \pm 1.30}$ & $83.89{\scriptstyle \pm 2.15}$ & $75.54{\scriptstyle \pm 1.32}$ & $\underline{64.74{\scriptstyle \pm 0.43}}$ & $48.80{\scriptstyle \pm 3.16}$ & $30.56{\scriptstyle \pm 3.86}$ \\ 
        {NBFnet} & $71.65 {\scriptstyle \pm 2.27}$ & $74.07 {\scriptstyle \pm 1.75}$ & $58.73 {\scriptstyle \pm 1.99}$ & OOM & OOM & $4.00 {\scriptstyle \pm 0.58}$ \\ 
        {Neo-GNN} & $80.42 {\scriptstyle \pm 1.31} $ & $84.67{\scriptstyle \pm 2.16}$ & $73.93{\scriptstyle \pm 1.19}$ & $57.52 {\scriptstyle \pm 0.37}$ & $\underline{49.13{\scriptstyle \pm 0.60}}$ & $63.57{\scriptstyle \pm 3.52}$ \\ 
        {BUDDY} & $\underline{88.00{\scriptstyle \pm 0.44}}$ & $\mathbf{92.93{\scriptstyle \pm 0.27}}$ & $74.10{\scriptstyle \pm 0.78}$ & $\mathbf{65.94{\scriptstyle \pm 0.58}}$ & $\mathbf{49.85{\scriptstyle \pm 0.20}}$ & $\mathbf{78.51{\scriptstyle \pm 1.36}}$ \\ 
\midrule
        ENGNN & $\mathbf{88.10_{\pm 1.67}}$ & $\underline{91.56_{\pm 1.02}}$ & $\mathbf{81.26_{\pm 1.20}}$ & $63.69_{\pm 0.82}$ & $44.97_{\pm 0.74}$ & $\underline{77.61_{\pm 4.50}}$ \\ 
        MPNN & $86.26_{\pm 1.64}$ & $90.40_{\pm 1.71}$ & $\underline{79.48_{\pm 3.74}}$ & $62.84 _{\pm 1.07}$ & $5.62_{\pm 2.52}$ & $24.76_{\pm 15.29}$ \\ 
        NMPNN & $48.12_{\pm 11.94}$ & $68.63_{\pm 7.29}$ & $63.96_{\pm 1.92}$ & $7.35_{\pm 7.04}$ & $39.90_{\pm 5.52}$ & $23.08_{\pm 5.89}$
         \\ \bottomrule
\end{tabular}
\vskip -0.15in
\end{table*}
\section{Related Work}

Using randomness to enhance the expressivity of Graph Neural Networks (GNNs) has become a common approach. Early methods, like \citet{GNN-RNI} and \citet{rGIN}, add noise to node features, while \citet{DropGNN} introduces noise by randomly dropping edges, all showing improvements in expressivity. However, these methods struggle with generalization due to the increased sample complexity from noise. \citet{mplp} and \citet{randomeigen} address this by using random noise in linear GNN layers and learning features that either estimate neighborhood overlap or converge to the graph’s Laplacian eigenvectors. While these methods make random features more meaningful, they still rely on task-specific constraints or linearity. In contrast, our approach addresses noise-induced generalization issues by processing it equivariantly, making it applicable to all graph tasks. We also unify previous methods by showing that \citet{mplp} achieve invariance to any orthogonal transformation of noise, and \citet{randomeigen} achieve invariance to transformations preserving the Laplacian eigenvectors, while our method generalizes these invariances for broader applicability.

\begin{table*}[t]
\centering
\setlength{\tabcolsep}{2pt}
\caption{Mean Micro-F1 with standard error of the mean. Results are provided from runs with ten random seeds.}\label{tab::subgraph}
\vskip 0.1in
\resizebox{\textwidth}{!}{
\begin{tabular}{lcccccccc}
\toprule
Method & density & cut ratio & coreness & component & ppi-bp & hpo-metab & hpo-neuro & em-user \\ \midrule
GLASS & $\underline{0.930_{\pm 0.009}}$ & $\underline{0.935_{\pm 0.006}}$ & $\mathbf{0.840_{\pm 0.009}}$ & $\mathbf{1.000_{\pm 0.000}}$ & $\mathbf{0.619_{\pm 0.007}}$ & $\mathbf{0.614_{\pm 0.005}}$ & $\mathbf{0.685_{\pm 0.005}}$ & ${0.888_{\pm 0.006}}$ \\
SubGNN & $0.919_{\pm 0.006}$ & $0.629_{\pm 0.013}$ & $0.659_{\pm 0.031}$ & $0.958_{\pm 0.032}$ & $\underline{0.599_{\pm 0.008}}$ & $0.537_{\pm 0.008}$ & $\underline{0.644_{\pm 0.006}}$ & $0.816_{\pm 0.013}$ \\
Sub2Vec & $0.459 _{\pm 0.012}$ & $0.354_{\pm 0.014}$ & $0.360_{\pm 0.019}$ & $0.657_{\pm 0.017}$& $0.388_{\pm 0.001}$ & $0.472_{\pm 0.010}$ & $0.618_{\pm 0.003}$ & $0.779_{\pm 0.013}$ \\
\midrule
ENGNN & $\mathbf{0.992_{\pm 0.003}}$ & $\mathbf{0.984_{\pm 0.007}}$ & $\underline{0.742_{\pm 0.014}}$ & $\mathbf{1.000_{\pm 0.000}}$ & $0.581_{\pm 0.007}$ & $\underline{0.540_{\pm 0.008}}$ & $0.590_{\pm 0.003}$ & $\mathbf{0.902_{\pm 0.006}}$\\
MPNN & $0.321_{\pm 0.023}$ & $0.311_{\pm 0.012}$&$0.545_{\pm 0.024}$ & $\mathbf{1.000_{\pm 0.000}}$ & $0.547_{\pm 0.009}$ & $ 0.500_{\pm 0.010}$ & $ 0.587_{\pm 0.004}$ & $ 0.641_{\pm 0.017}$\\
NMPNN & $0.321 _{\pm 0.023}$ & $0.311_{\pm 0.012 }$ &$0.527_{\pm 0.016}$ & $\mathbf{1.000_{\pm 0.000}}$ & $0.516_{\pm 0.010 }$ & $ 0.460_{\pm 0.010}$ &$ 0.582_{\pm 0.006}$ & $\underline{0.896_{\pm 0.006}}$\\
\bottomrule
\end{tabular}
}
\end{table*}
\begin{table*}
\vskip -0.1in
\caption{Normalized MAE results of counting substructures on synthetic dataset. The colored cell means an error less than 0.01.}\label{tab::count} 
\vskip 0.1in
\centering
\begin{small}
\begin{tabular}{lccccccccc}
\toprule
Method& 3-Cyc. & 4-Cyc. & 5-Cyc. & 6-Cyc. & Tail. Tri. & Chor. Cyc. & 4-Cliq. & 4-Path & Tri.-Rect.     \\ \midrule
MPNN & 0.3515 & 0.2742   &  0.2088   & 0.1555& 0.3631 & 0.3114 & 0.1645 & 0.1592 & 0.2979 \\
ID-GNN  & {\color{yellow}0.0006} & {\color{yellow}0.0022} &0.0490 & 0.0495  & 0.1053 & 0.0454 &{\color{yellow}0.0026} & 0.0273 & 0.0628        \\
NGNN    &{\color{yellow}0.0003}  & 0.0013 & 0.0402 & 0.0439  &0.1044 & 0.0392&{\color{yellow}0.0045} & 0.0244 & 0.0729   \\
GNNAK+  &{\color{yellow}0.0004}&{\color{yellow}0.0041}  &0.0133  &0.0238 &{\color{yellow}0.0043}&0.0112&{\color{yellow}0.0049}&{\color{yellow}0.0075} &0.1311  \\
PPGN  &  {\color{yellow}0.0003}    &   {\color{yellow}0.0009}  &  {\color{yellow}0.0036} &  {\color{yellow}0.0071}  & {\color{yellow}0.0026} &{\color{yellow}0.0015} & 0.1646 &{\color{yellow}0.0041} & 0.0144\\
I$^2$-GNN  & {\color{yellow}0.0003}  &  {\color{yellow}0.0016}   & {\color{yellow}0.0028} &  {\color{yellow}0.0082}  &{\color{yellow}0.0011} &{\color{yellow}0.0010} &{\color{yellow}0.0003} &{\color{yellow}0.0041} &{\color{yellow}0.0013} \\
DRFW2  & {\color{yellow}0.0004} & {\color{yellow}0.0015} & {\color{yellow}0.0034} & {\color{yellow}0.0087}& {\color{yellow}0.0030} &{\color{yellow}0.0026} &{\color{yellow}0.0009} &{\color{yellow}0.0081} &{\color{yellow}0.0070} \\
\midrule
    ENGNN & {\color{yellow}0.0030} & {\color{yellow}0.0047} & {\color{yellow}0.0058} & {\color{yellow}0.0078} & {\color{yellow}0.0038} & {\color{yellow}0.0031} &{\color{yellow}0.0016}& {\color{yellow}0.0033} &{\color{yellow}0.0065}\\
MPNN & 0.1960 & 0.1808 & 0.1658 & 0.1313 & 0.1585 & 0.1294 & 0.0598 & 0.0594 & 0.1400\\
NMPNN & {\color{yellow}0.0031} & 0.0121 & 0.0167 & 0.0228 & 0.0182 & 0.0179 & 0.0128 & 0.0168 & 0.0572\\
\bottomrule
\end{tabular}
\end{small}
\label{exp_count_cycle}
\end{table*}

\begin{table}[t]
\vskip -7pt
    \caption{Results on graph property prediction tasks.}\label{tab::zinc}
\vskip 7pt
    \centering
    \setlength{\tabcolsep}{1pt}
    \begin{tabular}{lccc}
    \toprule
        ~ & zinc & zinc-full & molhiv \\
     ~ & MAE$\downarrow$ & MAE$\downarrow$ & AUC$\uparrow$ \\
      \midrule
        GIN & $0.163_{\pm0.004}$ & $0.088_{\pm0.002}$ & $77.07_{\pm1.49}$ \\ 
        GNN-AK+ & $0.080_{\pm0.001}$ & – & $79.61_{\pm1.19}$ \\ 
        ESAN & $0.102_{\pm0.003}$ & $0.029_{\pm0.003}$ & $78.25_{\pm0.98}$ \\ 
        SUN & $0.083_{\pm0.003}$ & $0.024_{\pm0.003}$ & $80.03_{\pm0.55}$ \\ 
        SSWL & $0.083_{\pm0.003}$ & $0.022_{\pm0.002}$ & $79.58_{\pm0.35}$ \\ 
        DRFWL & $0.077_{\pm0.002}$ & $0.025_{\pm0.003}$ & $78.18_{\pm2.19}$ \\ 
        CIN & $0.079_{\pm0.006}$ & $0.022_{\pm0.002}$ & $80.94_{\pm0.57}$ \\ 
        NGNN & $0.111_{\pm0.003}$ & $0.029_{\pm0.001}$ & $78.34_{\pm1.86}$ \\
    \midrule
        ENGNN &$0.114_{\pm 0.005}$&$0.044_{\pm 0.002}$&$78.51_{\pm 0.86}$\\
        MPNN &$0.131_{\pm 0.007}$&$0.046_{\pm 0.002}$&$78.27_{\pm 1.14}$\\
        NMPNN &$0.136_{\pm 0.007}$&$0.051_{\pm 0.004}$&$77.74_{\pm 0.98}$\\
    \bottomrule
    \end{tabular}
    \vskip -10pt
\end{table}
\section{Experiments}

In this section, we extensively evaluate the performance of our Equivariant Noise MPNN (EN-MPNN) on node, link, subgraph, and whole graph tasks. For ablation, we also evaluate vanilla MPNN with noise as input (NMPNN) and ENGNN replacing noise with zero as input (MPNN). As previous work injecting noise to vanilla GNN~\citep{rGIN, GNN-RNI} is primarily implemented on synthetic datasets, we directly use NMPNN as baseline. Detailed experimental settings are included in Appendix~\ref{app:exp}. The statistics and splits of datasets are shown in Appendix~\ref{app:data}. 

\subsection{Node Tasks}

We evaluate our models on real-world node classification tasks: For homogeneous graphs, we include three citation graph datasets, Cora, CiteSeer and PubMed~\citep{Cora}, and two Amazon co-purchase graphs, Computers and Photo~\citep{Photo}. We also use heterogeneous graphs, including Wikipedia graphs Chameleon and Squirrel~\citep{Chameleon}, the Actor co-occurrence graph from WebKB3~\citep{Texas}. We randomly split the node set into train/validation/test sets with a ratio of $60\%/20\%/20\%$. Our models are compared with widely used node classification GNNs: GCN~\citep{GCN}, APPNP~\citep{APPNP}, ChebyNet~\citep{ChebNet}, GPRGNN~\citep{GPRGNN}, and BernNet~\citep{BernNet}.

The experimental results are presented in Table~\ref{tab::node}. ENGNN surpasses all baselines on 6/8 datasets. Moreover, for ablation study, adding noise feature to vanilla MPNN leads to significant performance decrease as NMPNN decrease accuracy by $32\%$ compared with MPNN on average, verifying that noise feature can leads to bad generalization. However, ENGNN outperforms MPNN and NMPNN consistently on all datasets and achieves $1\%$ and $33\%$ accuracy gain on average, illustrating that utilizing noise can boost expressivity and equivariance can alleviate the large sample complexity problem caused by noise feature.
\subsection{Link}
We evaluate our models on link prediction datasets, including three citation graphs~\citep{Cora} (Cora, Citeseer, and Pubmed) and three Open Graph Benchmark~\citep{OGB} datasets (Collab, PPA, and DDI). We employ a range of baseline methods, encompassing traditional heuristics like CN~\citep{CommonNeighbor}, RA~\citep{RA}, and AA~\citep{AA}, as well as GAE models, such as GCN~\citep{GCN} and SAGE~\citep{GraphSage}. Additionally, we consider SF-then-MPNN models, including SEAL~\citep{SEAL} and NBFNet~\citep{NBFNet}, as well as SF-and-MPNN models like Neo-GNN~\citep{Neo-GNN} and BUDDY~\citep{Gsketch}. The baseline results are sourced from~\citep{Gsketch}. 

The experimental results are presented in Table~\ref{tab::link}. Our ENGNN achieves best performance on 2 datasets and second best performance on 2 datasets. It also achieves comparable performance on other datasets. Compared with MPNN and NMPNN , ENGNN outperforms them consistently on all datasets and achieves $16\%$ and $33\%$ score gain on average, respectively, illustrating that utilizing noise can boost expressivity and equivariance can alleviate the large sample complexity problem caused by noise feature.

\subsection{Subgraph}

We evaluate our models on subgraph classfication tasks. Datasets include four synthetic datasets: \texttt{density}, \texttt{cut ratio}, \texttt{coreness}, \texttt{component}, and four real-world subgraph datasets, namely \texttt{ppi-bp}, \texttt{em-user}, \texttt{hpo-metab}, \texttt{hpo-neuro}~\citep{SubGNN}.  We consider three baseline methods: \textbf{SubGNN}~\citep{SubGNN} with subgraph-level message passing,  \textbf{Sub2Vec}~\citep{Sub2Vec} sampling random walks in subgraphs and encoding them with RNN, \textbf{GLASS}~\citep{GLASS} using MPNN with labeling trick. 

The results are shown in Table~\ref{tab::subgraph}, our ENGNN achieves best performance on 4/8 datasets. Moreover, On all datasets, ENGNN consistently outperforms MPNN and NMPNN by a large margin, illustrating the expressivity gain provided by noise and the generalization provided by symmetry.

\subsection{Graph}

For graph-level tasks, we evaluate our model on subgraph counting task and some molecule datasets. Baseline includes various expressive high order GNNs~\citep{IDGNN, NGNN, GNNAK, I2GNN, PPGN, DRFWL}. First, on subgraph counting dataset, we follow the setting of \citet{DRFWL}, where a model achieves loss lower than $0.01$ can count the corresponding subgraph. The results is shown in Table~\ref{tab::count}. The subgraphs include 3-6-Cyc (cycle of length from 3 to 6), Tail. Tri. (Tailed Triangle), Chor. Cyc. (cycle with a chord), 4-Cliq (4-Clique), 4-Path (path of length 4), and Tri-Rect. (Triangle connected with an rectangle). While vanilla MPNN cannot count any of these subgraphs, ENGNN can count all of them. Note that though NMPNN does not achieve error lower than $0.01$, it still fit much better than MPNN, illustrating the expressivity gain by introducing noise.

The results on molecule datasets are shown in Table~\ref{tab::zinc}. ENGNN's performance is not favorable compared with other models. Partially because these datasets are saturated and have many specific design like positional encoding and motif extracting that ENGNN does not use. Moreover, our ENGNN has a much smaller time and space complexity as shown in Table~\ref{tab:scalability}. ENGNN still achieves performance gain compared with MPNN and NMPNN.

\subsection{Scalability}

We present training time per epoch and GPU memory consumption data in Table~\ref{tab:scalability}. Our ENGNN keeps MPNN architecture and consumer significant less resource than existing expressive GNN models.

\begin{table}[t]
\vskip - 0.1 in
\caption{Training time per epoch and GPU memory consumption on zinc dataset with batch size 128.}\label{tab:scalability}
\vskip 0.1 in
    \centering
    \setlength{\tabcolsep}{2pt}
    \begin{small}    
    \begin{tabular}{lccccc}
    \toprule
        ~ & MPNN & ENGNN &SUN & SSWL & PPGN \\ 
    \midrule
        Time/s & 2.36& 4.81& 20.93  & 45.30  & 20.21   \\ 
        Memory/GB& 0.24 & 0.62 & 3.72  & 3.89  & 20.37    \\ 
    \bottomrule
    \end{tabular}
    \end{small}
\vskip -0.1 in
\end{table}

\section{Conclusion}
We analyse the generalization bounds of GNN with noise as auxliary input. Additionally, we propose equivariant noise GNN, a GNN that utilize noise equivariantly for better generalization bound. Our approach demonstrates universal theoretical expressivity and excels in real-world performance. It extends the design space of GNN and provides a principled way to utilize noise feature.

\section{Limitations}
Although the ENGNN introduced in our work shares the same time complexity as traditional MPNNs, the inclusion of noise features introduces additional computational overhead. Furthermore, despite the fact that noise is not task-specific, our approach requires modifications to the aggregator, which means it cannot be seamlessly integrated with other existing GNNs. Future work will focus on further reducing computational complexity and developing a more plug-and-play method.

\section*{Impact Statement}

This paper presents work whose goal is to advance the field of graph representation learning and will improve the design of graph generation and prediction models. There are many potential societal consequences of graph learning improvement, such as accelerating drug discovery, improving traffic efficiency, and providing better recommendation in social network. None of them we feel need to be specifically highlighted here for potential risk.
\newpage
\bibliography{example_paper}
\bibliographystyle{icml2025}

\newpage
\appendix
\onecolumn

\section{Proof for Generalization Bound}\label{app:gen_proof}

We first restate our problem setting. 

Consider a graph $ G = (A, x) \in \mathcal{G} $ with $ n $ nodes, where $ A \in \mathbb{R}^{n \times n} $ is the adjacency matrix and $ x \in \mathbb{R}^{n \times d} $ is the node feature matrix. Additionally, there is noise $ Z \in \mathcal{Z} = \mathbb{R}^{n \times C} $. We introduce a group $ S_n \times T $, where $S_n$ is $n$-order permutation group. For a permutation $ \pi_1\in S_n$ and an operation $t\in T$ on the noise, the group acts on the data $ (A, X, Z) $ as $ (\pi(A), \pi(X), \pi(t(Z))) $, where $ \pi(A) \in \mathbb{R}^{n \times n} $, $ \pi(A)_{\pi(i), \pi(j)} = A_{ij} $, $ \pi(X) \in \mathbb{R}^{n \times d} $, $ \pi(X)_{\pi(i)} = X_i $, and $ \pi(t(Z)) \in \mathbb{R}^{n \times C} $, with $T$ being an operation that commutes with any permutation. We require $T$ to be a compact group. The input domain is $ \mathcal{X} = \mathcal{G} \times \mathcal{Z} $ and the target domain is $\mathcal{Y}$. Let $ \mathcal{W} = \mathcal{X} \times \mathcal{Y} $ represent the set of data points. A learning task is defined as $(X, Y, l) $, where $ X $ and $ Y $ are random elements in $ \mathcal{X} $ and $\mathcal{Y}$, respectively, and $ l: Y \times Y \to \mathbb{R}^+ $ is an integrable loss function.

The PAC framework provides a foundation for learning from data. Let $ H $ be a class of measurable functions $ h: \mathcal{X} \to \mathcal{Y} $, known as the hypothesis class. An algorithm $ \text{alg}: \bigcup_{i \in \mathbb{N}} \mathcal{W}^i \to H $ maps a finite sequence of data points to a hypothesis in $ H $. We say that $ \text{alg} $ learns $ H $ with respect to a task $ (X, Y, l) $ if there exists a function $ m: (0, 1)^2 \to \mathbb{N} $ such that for all $ \epsilon, \delta \in (0, 1) $, if $ n > m(\epsilon, \delta) $, then:
\begin{equation}
P \left( \mathbb E[l(h_S(X), Y) \mid S] \ge \inf_{h \in H} \mathbb E[l(h(X), Y)] + \epsilon \right) \ge 1 - \delta,
\end{equation}
where $ h_S = \text{alg}(S)$ and $ S \sim (X, Y)^n $ is an i.i.d. sample. The sample complexity of $ \text{alg} $ is the minimum $ m(\epsilon, \delta) $ that satisfies this condition.

We first restate the basic concentration bound:

\begin{equation}
P[\sup_{h\in \mathcal H}| E[l(h(X), Y)]-\frac{1}{n}\sum_{i=1}^n l(h(X_i), Y_i)|\ge \epsilon]
\le 2 \inf Conv
\end{equation}

\begin{proof}
Given random variable $X, Y$, define $L(h)= E[l(h(X), Y)]-\frac{1}{n}\sum_{i=1}^n l(h(X_i), Y_i)$.

Therefore, given $h, h'\in \mathcal H$,

\begin{align}
    |L(h)-L(h')| &= |E[l(h(X), Y)-l(h'(X), Y)] - \frac{1}{n} \sum_{i=1}^n [l(h(X_i), Y_i)-l(h'(X_i), Y_i)]|\\
    &\le E[|l(h(X), Y)- l(h'(X), Y)|]+\frac{1}{n} \sum_{i=1}^n |l(h(X_i), Y_i)-l(h'(X_i), Y_i)| \\
    &\le 2C_l\Vert h-h'\Vert
\end{align}
where $\Vert h-h'\Vert=\sup_{x\in\mathcal X} \Vert h(X)-h'(X)\Vert$, $C_l$ is the Lipschitz constant of $l$.  

Let $K$ be an $\kappa$-cover of $\mathcal H$. Define set $D(k)=\{h\in \mathcal H| \Vert h-k\Vert \le \kappa\}$. Then

\begin{align}
    P[\sup_{h\in H} |L(h)|\ge \epsilon]&=P[\sup_{k\in K}\sup_{h\in D(k)} |L(h)|\ge\epsilon]\\
    &\le \sum_{k\in K} P[\sup_{h\in D(k)}|L(h)|\ge\epsilon]\\
    &\le \sum_{k\in K} P[|L(k)|+2C_l\kappa\ge\epsilon]\\
    &\le \sum_{k\in K} P[|L(k)|\ge (1-\alpha)\epsilon]
\end{align}
where we define $\alpha=\frac{2C_l\kappa}{\epsilon}$.

\begin{align}
    P[|L(k)|\ge (1-\alpha) \epsilon]&=P[|E[\frac{1}{n}\sum_{i=1}^nl(h(X_i), Y_i)]-\frac{1}{n}\sum_{i=1}^n l(h(X_i), Y_i)|\ge (1-\alpha) \epsilon]\\
    &=2\exp{(-2n(1-\alpha)^2\epsilon^2)}.
\end{align}
The second step is Hoeffding's Inequality and $l(h(X), Y)\in [0, 1]$. Therefore, for all $\alpha\in (0, 1)$
\begin{align}
     P[\sup_{h\in H} |L(h)|\ge \epsilon]&\le 2|K| \exp{(-2n(1-\alpha)^2\epsilon^2)}\\
     &\le 2N(\mathcal H, \Vert \cdot\Vert, \frac{\alpha\epsilon}{2C_l}) \exp{(-2n(1-\alpha)^2\epsilon^2)}
\end{align}
With $\alpha=\frac{1}{2}$, 
\begin{equation}
    P[\sup_{h\in H} |L(h)|\ge \epsilon]\le 2N(\mathcal H, \Vert \cdot\Vert, \frac{\epsilon}{4C_l}) \exp{(-\frac{n\epsilon^2}{2})}
\end{equation}
\end{proof}
Then the sample complexity is $\Omega(\frac{1}{\epsilon^2}(\ln N(\mathcal H, \Vert\cdot\Vert, \frac{\epsilon}{4C_l})+\ln \frac{1}{\delta}))$

Then we discretize functions to vector to bound the covering number of $\mathcal H$ by covering number of $\mathcal {X}$ and $\mathcal{Y}$.

\begin{proof}
Given an $\delta_1$-cover $I$ of $X$, $\delta_2$-cover $J$ of $Y$. We build a set of function as follows.
\begin{equation}
    F=\{f_{j}| j\in J^{|I|}, f_j(x)=j_{l} \text{if } x\in I_l\}.
\end{equation}
Therefore, for all $h\in\mathcal H$.
\begin{align}
    \min_{f\in F}\Vert f-h\Vert 
    & = \min_{f\in F}\sup_{X\in \mathcal X}\Vert f(X)- h(X)\Vert\\
    & = \min_{f\in F}\max_{i\in I} \sup_{x\in D(i)}\Vert f(X)- h(X)\Vert\\
    & \le \min_{f\in F}\max_{i\in I} \sup_{x\in D(i)} \Vert f(x)-f(i)\Vert+\Vert h(x)-h(i)\Vert+\Vert f(i)- h(i)\Vert\\
    & \le \min_{f\in F}\max_{i\in I} C\delta_1+\Vert f(i)- h(i)\Vert\\
    & \le  C\delta_1+\delta_2\\
\end{align}

Therefore, $F$ is an $C\delta_1+\delta_2$-cover of $\mathcal H$. Therefore, with $\delta_1=\frac{\delta}{2C}$ and $\delta_2=\frac{\delta}{2}$.
\begin{equation}
    N(\mathcal H, \Vert \cdot \Vert, \delta)
    \le N(\mathcal Y, \Vert \cdot \Vert, \frac{\delta}{2})^{N(\mathcal X, \Vert \cdot \Vert, \frac{\delta}{2C})}
\end{equation}
\end{proof}

Then, the sample complexity can be transformed to 
\begin{equation}
    \Omega(\frac{1}{\epsilon^2}(\ln\frac{1}{\delta}+N(\mathcal X, \rho, \frac{\delta}{2C})\ln N(\mathcal Y, \Vert\cdot\Vert, \frac{\delta}{2})))
\end{equation}

When separating the input space $\mathcal X$ into $\mathcal G\times \mathcal Z$, and let the hypothesis space be functions partial lipschitz to $\mathcal G$ with lipschitz constant $C_G$ and to $\mathcal Z$ with lipschitz constant Z, the covering number of $\mathcal H$ then becomes.

\begin{proof}
Given an $\delta_1$-cover $I$ of $\mathcal G$, $\delta_2$-cover $J$ of $\mathcal Z$, and $\delta_3$-cover $K$ of $Y$. We build a set of function as follows.
\begin{equation}
    F=\{f_{k}| k\in K^{|I|\times |J|}, f_k(G, Z)=k_{ij} \text{if } G\in D(I_i), Z\in D(J_j)\}.
\end{equation}
Therefore, for all $h\in\mathcal H$.
\begin{align}
    \min_{f\in F}\Vert f-h\Vert 
    & = \min_{f\in F}\sup_{X\in \mathcal X}\Vert f(X)- h(X)\Vert\\
    & = \min_{f\in F}\max_{i\in I}\max{j\in J} \sup_{G\in D(i)}\sup_{Z\in D(j)}\Vert f(G, Z)- h(G, Z)\Vert\\
    & \le \min_{f\in F}\max_{i\in I}\max{j\in J} \sup_{G\in D(i)}\sup_{Z\in D(j)}  \Vert f(G, Z)-f(i, Z)\Vert+\Vert f(i, Z)-f(i, j)\Vert+\Vert f(i,j)- h(i,j)\Vert\\
    & \le \min_{f\in F}\max_{i\in I} \max_{j\in Z} C_G\delta_1+C_Z\delta_2+\Vert f(i, j)- h(i, j)\Vert\\
    & \le  C_G\delta_1+C_Z\delta_2+\delta_3\\
\end{align}

Therefore, $F$ is an $C_G\delta_1+C_Z\delta_2+\delta_3$-cover of $\mathcal H$. Therefore, with $\delta_1=\frac{\delta}{3C_G}, \delta_2=\frac{\delta}{3C_Z}$ and $\delta_3=\frac{\delta}{3}$.
\begin{equation}
    N(\mathcal H, \Vert \cdot \Vert, \delta)
    \le N(\mathcal Y, \Vert \cdot \Vert, \frac{\delta}{3})^{N(\mathcal G, \rho_G, \frac{\delta}{3C_G})N(\mathcal Z, \rho_Z, \frac{\delta}{3C_Z})}
\end{equation}
\end{proof}

Then, the sample complexity can be transformed to 
\begin{equation}
    \Omega(\frac{1}{\epsilon^2}(\ln\frac{1}{\delta}+N(\mathcal G, \rho_G, \frac{\delta}{3C_G})N(\mathcal Z, \rho_Z, \frac{\delta}{3C_Z})\ln N(\mathcal Y, \Vert\cdot\Vert, \frac{\delta}{3})))
\end{equation}

\section{Proof for Expressivity}\label{app:exp_proof}
\subsection{Proof of Proposition~\ref{prop:aggr_equiv}}
Note that DeepSet model is permutation invariant to permutation on the dimension it aggregates. Moreover, if operator act individually on some dimension, the operator is also equivariant to permutation on the dimension.

Therefore, when $Z\to P_2(P_1(Z))$, $\psi(Z_{:,:,k})=\psi(Z_{:,:,P_2^{-1}(k)})$, so $Z^1\to P_2(P_1(Z^1))$.

With $Z^1\to P_2(P_1(Z^1))$, and $x\to P_1(x)$, $x^{0}\to P_1(x^{0})$.

$X^1\to X^1$.

$X^2\to P_1(X^2)$

$Z^2\to P_2(P_1(Z^2))$.

\subsection{Proof of Proposition~\ref{prop:aggr_exp}}

Following \citet{NewDeepSet}, we first show that the set encoding $X^1$ in our aggregator can approximate any invariant function first. As permutation equivariant function can be expressed as a elementwise transformation conditioned by the invariant function, we can easily approximate any equivariant output.  

According to Stone-Weierstrass theorem, to prove the universality of set encoding $X^1$ is equivalent to that our aggregator can differentiate any two input set of invariant and equivariant features with some parameterization. 

Let all DeepSet and MLPs in our aggregator be injective. Given two set of features $X\in \mathbb{R}^{k\times d}, Z\in \mathbb{R}^{k\times L\times C}$ and $X'\in \mathbb{R}^{k\times d}, Z'\in \mathbb{R}^{k\times L\times C}$, if $X^1={X'}^1$, then,

\begin{itemize}
    \item As $\varphi$ is injective, $\exists P_1\in S_k$, $P_1(X^0)={X'}^0$.
    \item As $P_1(X^0)={X'}^0$, $P_1(X)=X'$. Moreover, $P_1(\phi(Z^1))=\phi({Z'}^1)$.
    \item As $P_1(\phi(Z^1))=\phi({Z'}^1)$, for each row $\phi(Z^1)_{i}=\phi({Z'}^1)_{P_1(i)}$, $\exists P_{2i}\in S_C$, $Z^1_i=P_{2i}({Z'}^1_{P_1(i)})$. The permutation of noise channel may be different for each row, but each noise channel is assigned with unique column label, so $P_{2i}$ are all equal to $P_2$. Therefore, $P_2(P_1(Z^1))={Z'}^1$.
    \item So $P_2(P_1(Z))=Z'$,
\end{itemize}
Therefore, the invariant representation is universal. 

\subsection{Proof of Theorem~\ref{thm:graph_exp}}
Given two graphs $G$ and $H$, such that $\exists P\in S_n, G=P(H)$. Let $\text{ENGNN}$ denote our model with both graph and noise as input. Our model is invariant to node permutation of input.
\begin{align}
    \mathbb{E}_{Z\in \mathcal{Z}}\text{ENGNN}(G, Z)
    &=\mathbb{E}_{Z\in \mathcal{Z}}\text{ENGNN}(P(H), P(P^{-1}(Z)))\\
    &=\mathbb{E}_{Z\in \mathcal{Z}}\text{ENGNN}(P(H), P(P^{-1}(Z)))\\
    &=\mathbb{E}_{Z\in \mathcal{Z}}\text{ENGNN}(H, P^{-1}(Z))\\
\end{align}
As noise distribution for $Z$ is invariant to node permutation,
\begin{align}
    \mathbb{E}_{Z\in \mathcal{Z}}\text{ENGNN}(G, Z)
    &=\mathbb{E}_{Z\in \mathcal{Z}}\text{ENGNN}(H, Z).
\end{align}

Let $A\to B$ denote there exists an injective mapping from $A$ to $B$, where $A$, $B$ are representations of input graph and noise. For graph $G$, given noise $Z$, each node $i$'s invariant representation $u_i$ and equivariant representation $v_i$ contains neighbor edge connectivity information,
\begin{equation}
    u_i, v_i\xrightarrow{} X_i, Z_i, \{\{(X_j, Z_j)|(i,j)\in E\}\},
\end{equation}
Therefore, pooling all nodes' representations leads to total edge connectivity:
\begin{equation}
    \text{AGGR}(\{\{(u_i,v_i)|i\in V\}\}) \to \{\{(Z_i, Z_j)|(i,j)\in E\}\}, \{(x_i, Z_i)|i\in V\},
\end{equation}
as input noise $Z_i$ are unique for each node, the output representation can determine the input node features and edges. Therefore, if two graphs are not isomorphic, they produce different representations with noise.

\subsection{Proof of Theorem~\ref{thm:subgraph_exp}}
Given two graphs with node subsets $(G, U_G)$ and $(H, U_H)$, such that $\exists P\in S_n, G=P(H), U_G=P(U_H)$. 
\begin{align}
    \mathbb{E}_{Z\in \mathcal{Z}}\text{ENGNN}(G, U_G, Z)
    &=\mathbb{E}_{Z\in \mathcal{Z}}\text{ENGNN}(P(H), P(U_H), P(P^{-1}(Z))\\
    &=\mathbb{E}_{Z\in \mathcal{Z}}\text{ENGNN}(H, U_H, P^{-1}(Z))\\
\end{align}
As noise distribution for $Z$ is invariant to node permutation,
\begin{align}
    \mathbb{E}_{Z\in \mathcal{Z}}\text{ENGNN}(G, U_G, Z)&=\mathbb{E}_{Z\in \mathcal{Z}}\text{ENGNN}(H, U_H, Z).
\end{align}

Let $A\to B$ denote there exists an injective mapping from $A$ to $B$, where $A$, $B$ are representations of input graph and noise. For graph $G$ with node set $U_G$, given noise $Z$, each node $i$'s invariant representation $u_i$ and equivariant representation $v_i$ contains neighbor edge connectivity information,
\begin{equation}
    u_i, v_i\xrightarrow{} X_i, Z_i, \{\{(X_j, Z_j)|(i,j)\in E\}\},
\end{equation}
Therefore, pooling all nodes' representations leads to total edge connectivity:
\begin{equation}
    \text{AGGR}(\{\{(u_i,v_i)|i\in V\}\}) \to \{\{(Z_i, Z_j)|(i,j)\in E\}\}, \{(x_i, Z_i)|i\in V\}
\end{equation}
\begin{equation}
    \text{AGGR}(\{\{(u_i,v_i)|i\in U_G\}\}) \to  \{(Z_i)|i\in U_G\}.
\end{equation}
Therefore, the resulting representation
\begin{equation}
    \mathbb{E}_{Z\in \mathcal{Z}}\text{ENGNN}(G, U_G, Z)\to \{(Z_i)|i\in U_G\},\{\{(Z_i, Z_j)|(i,j)\in E\}\}, \{(x_i, Z_i)|i\in V\}
\end{equation}
as input noise $Z_i$ are unique for each node, the output representation can determine the input node features, edges, and node subset. Therefore, if two graphs with node subsets are not isomorphic, they produce different representations with noise.
\section{Experimental Setting}\label{app:exp}

We use PyTorch and PyTorch Geometric for model development. All experiments are conducted on an Nvidia 4090 GPU on a Linux server. We use the AdamW optimizer with a cosine annealing scheduler. We use L1 loss for regression tasks and cross-entropy loss for classification tasks. 

We perform random search using Optuna to optimize hyperparameters by maximize valid score. The selected hyperparameters for each model are available in our code. The hyperparameter we tune include number of layer in [2, 10], hidden dimention in [32, 128], number of noise channel in [16, 64], number of noise feature dimension in [16, 64], learning rate in [1e-4, 1e-2], weight decay in [1e-6, 1e-1]. 

For baselines, we directly use the score reported in the original paper.

\section{Dataset}\label{app:data}

\begin{table}[t]
    \centering
    \caption{Statistics of the datasets. \#Nodes and \#Edges denote the number of nodes and edges per graph. In split column, 'fixed' means the dataset uses the split provided in the original release. 
 Otherwise, it is of the formal training set ratio/valid ratio/test ratio.}
    \label{tab::graphdata}
    \begin{small}
    \begin{tabular}{ccccccc}
    \toprule
        ~ & \#Graphs &  \#Nodes &  \#Edges & Task & Metric & Split \\ 
    \midrule
        Synthetic & 5,000 & 18.8  & 31.3  & Node Regression & MAE & 0.3/0.2/0.5. \\ 
        ZINC & 12,000 & 23.2  & 24.9  &  Regression & MAE & fixed \\ 
        ZINC-full & 249,456 & 23.2  & 24.9  &  Regression & MAE & fixed \\ 
        ogbg-molhiv & 41,127 & 25.5  & 27.5  &  Binary classification & AUC & fixed \\ 
    \bottomrule
    \end{tabular}
    \end{small}
\end{table}
For graph datasets, we summarize the statistics of our datasets in Table~\ref{tab::graphdata}. Synthetic is the dataset used in substructure counting tasks provided by ~\citet{I2GNN}, they are random graph with the count of substructure as node label. ZINC, ZINC-FULL~\citep{ZINC}, and ogbg-molhiv are three datasets of molecules. Ogbg-molhiv is one of Open Graph Benchmark dataset, which aims to use graph structure to predict whether a molecule can inhibits HIV virus replication. 

For subgraph datasets, we use the code provided by SubGNN to produce synthetic datasets and use the real-world datasets provided by SubGNN directly. The statistics of these datasets can be found in Table~\ref{tab:subgdatasets_detail}. As for dataset division, the real-world datasets take an 80:10:10 split, and the synthetic datasets follow a 50:25:25 split, following \citep{SubGNN}.

\begin{table}[t]
\centering
\caption{Detail of subgraph datasets}\label{tab:subgdatasets_detail}
\begin{tabular}{cccc}
\hline
Datasets & $|\sV|$ & $|\mathbb{E}|$ & Number of Subgraphs \\ \hline
density&5,000&29,521&250\\
cut-ratio&5,000&83,969&250\\
coreness&5,000&118,785&221\\
component&19,555&43,701&250\\
ppi\_bp&17,080&316,951&1,591\\
hpo\_metab&14,587&3,238,174&2,400\\
hpo\_neuro&14,587&3,238,174&4,000\\
em\_user&57,333&4,573,417&324
 \\\hline
\end{tabular}
\end{table}

\begin{table*}[t]
    \centering
    \caption{Statistics of link prediction dataset.}\label{tab:linkdatasets}
    \vskip 0.15in
    \begin{tabular}{l ccccccc}
    \toprule 
         &
         \textbf{Cora} &  
         \textbf{Citeseer} & 
         \textbf{Pubmed} &
         \textbf{Collab} &
         \textbf{PPA} &
         \textbf{DDI} &
         \textbf{Citation2}
        \\
\midrule
         
                  \#Nodes &

         2,708 & 
         3,327 &
         18,717 &
         235,868 &
         576,289 &
         4,267 &
         2,927,963
          \\
         
         \#Edges &
         5,278 & 
         4,676 &
         44,327 &
         1,285,465 &
         30,326,273 &
         1,334,889 &
         30,561,187
          \\

         splits &

         random &
         random & 
         random &
         fixed &
         fixed &
         fixed &
         fixed \\
          
         average degree &
         3.9 &
         2.74 & 
         4.5 &
         5.45 &
         52.62 &
         312.84 &
         10.44
        \\
         
         \bottomrule
\end{tabular}

\label{tab:subgraph properties}
\end{table*}
For link prediction datasets, random splits use $70\%/10\%/20\%$ edges for training/validation/test set respectively. Different from others, the collab dataset allows using validation edges as input on test set. We summarize the statistics of link datasets in Table~\ref{tab:linkdatasets}.

\begin{table}[h]
\centering
\caption{Node dataset statistics. }\label{tab::nodedatasets}
\vskip 0.15in
\begin{center}
\begin{small}
\begin{sc}
\setlength{\tabcolsep}{1mm}
{\begin{tabular}{ccccccccccc}
\hline
Datasets & Cora & CiteSeer & PubMed & Computers & Photo  & Chameleon & Squirrel & Actor & Texas & Cornell \\
\hline
$|V|$ & 2708 & 3327 & 19717 & 13752 & 7650 & 2277 & 5201 & 7600 & 183 & 183     \\
$|E|$ & 5278 & 4552 & 44324 & 245861 & 119081 & 31371 & 198353& 26659 & 279 & 277     \\
\hline
\end{tabular}}
\end{sc}
\end{small}
\end{center}
\vskip -0.1in
\end{table}

For node classification datasets, random splits use $60\%/20\%/20\%$ nodes for training/validation/test set respectively. We summarize the statistics of link datasets in Table~\ref{tab::nodedatasets}.

\end{document}